\documentclass{article}

\usepackage[preprint]{neurips_2026}
\usepackage{algorithm}
\usepackage{algpseudocode}
\usepackage{multirow}
\usepackage{multicol}
\usepackage{amsmath}
\usepackage{graphicx}
\usepackage{amsfonts}
\usepackage{subcaption}
\usepackage{booktabs}
\usepackage{multirow}
\usepackage{graphicx}

\usepackage[utf8]{inputenc} 
\usepackage[T1]{fontenc}    
\usepackage{hyperref}       
\usepackage{url}            
\usepackage{booktabs}       
\usepackage{amsfonts}       
\usepackage{subcaption,booktabs,multirow,graphicx,colortbl,xcolor}
\title{\textsc{Goal}: \underline{G}raph-based \underline{O}bjective-\underline{A}ligned Diffusion So\underline{l}vers for Dynamic Multi-Objective Optimization}

\author{%
  Xingyu Li\thanks{For correspondence.} \\
  School of Engineering Technology \\
  Purdue University \\
  West Lafayette, IN 47907 \\
  \texttt{li4558@purdue.edu} \\
}

\begin{document}

\maketitle

\begin{abstract} Existing neural combinatorial optimization solvers frame solution search as imitation of optimal decisions, inherently limiting their utility to single-objective minimization and static constraints. We propose \textsc{GOAL}, a conditioned diffusion solver over relational graph representations that enables controllable decision generations by conditioning on human-specified objectives. We introduce a heterogeneous graph encoding in which distinct edge types, corresponding to different classes of constraints, define the message passing structure of the graph neural network, which allows information to propagate selectively according to the ontology of each constraint. \textsc{GOAL} is instantiated and evaluated on three canonical scheduling benchmarks of various constraint complexity: the Flow Shop Problem (FSP), the Job Shop Scheduling Problem (JSP), and the Flexible Job Shop Scheduling Problem (FJSP). Generalization is demonstrated across structurally distinct constraint regimes and problem types without architectural modification. On all three benchmarks, \textsc{GOAL} achieves 100\% solution feasibility and near-zero MAPE (below 0.20\%) on multiple objectives for problem sizes up to 20 jobs and 60 operations, outperforming NSGA-II and MOEA/D in both solution quality and inference speed by up to 25× on $\epsilon$-feasible decision. \end{abstract}

\section{Introduction}

The rapid proliferation of autonomous assets, facilitated by advances in embodied AI~\citep{ha2023embodied, reed2022generalist} and physical AI~\citep{kaufmann2023champion, rudin2022learning}, has fundamentally transformed modern production and logistics systems. Large-scale deployments, such as Amazon fulfillment centers operating fleets of over one million autonomous robots~\citep{amazon2023}, exemplify a broader trend in which autonomous assets executing heterogeneous operations are progressively integrated into production workflows~\citep{lasi2014industry, brettel2014virtualization}. Despite significant opportunities to improve system flexibility and efficiency, this integration introduces new challenges in decision-making for dynamic scheduling and large-scale coordination of autonomous assets~\citep{yu2022surprising, gronauer2022multi}. For example, fixed machine assignments give way to dynamic task allocation~\citep{liu2023dynamic}, and static job sequences are disrupted by real-time order insertions and flexible rerouting~\citep{park2021schedulenet}. Furthermore, single-objective criteria such as makespan minimization~\citep{corsini2024self} prove insufficient to capture the full range of objectives within the operational environments with autonomy, including resilience, throughput balance, and, most critically, the preservation of human authority and trust~\citep{rahwan2019society, hadfield2016cooperative, li2026generativemanufacturing}.

Formally, such settings give rise to \textit{dynamic multi-objective optimization problems} (DMOPs), characterized by time-varying Pareto-optimal sets and fronts that require solvers to continuously adapt to dynamic environmental changes~\citep{jiang2022evolutionary, deb2002fast, li2008multiobjective}. The dynamism in decision-making for autonomy manifests along two distinct axes: changes in the \textit{objective function}, reflecting evolving human preferences over competing criteria such as makespan and schedule resilience~\citep{pinedo2016scheduling}, and, more critically, changes in the \textit{constraint topology}~\citep{branke2001evolutionary}. For instance, in the Job Shop Scheduling Problem (\textsc{JSP}), the introduction of humanoid robots capable of handling multiple operation types transforms the problem into a Flexible Job Shop Scheduling Problem (\textsc{FJSP})~\citep{brandimarte1993routing}, substantially expanding the feasible space. Simultaneously, real-time robot rerouting, dynamic task reallocation, and human-robot collaborative interactions introduce evolving precedence, assignment, synchronization, and capacity constraints, further increasing the combinatorial complexity of the decision-making \cite{hu2024multistage}.

Existing approaches for DMOPs, particularly in applications to the \textsc{JSP} and its variants, the Flow Shop Scheduling Problem (\textsc{FSP}) and \textsc{FJSP}, fall broadly into three categories, including dispatching heuristics, evolutionary methods, and neural combinatorial optimization. Dispatching heuristics, including Longest/Shortest Processing Time (LPT/SPT), Most/Least Work Remaining (MWKR/LWKR)~\citep{sels2012comparison, zhang2019review}, Longest Alternate Remaining Processing Time (LARPT)~\citep{pinedo2016scheduling}, and Dense Schedule/Longest Total Remaining Processing (DS/LTPR)~\citep{liaw1998iterative}, offer computational efficiency but optimize myopically for makespan minimization, with no mechanism for targeting a specific objective value. Evolutionary algorithms such as NSGA-II~\citep{deb2002fast} and MOEA/D~\citep{zhang2007moea} construct Pareto fronts through population-based search, enabling solution selection across a diverse set of objective trade-offs and constraints. However, both methods are fundamentally limited by their inability to transfer solutions across shifting objectives and constraints without full population reinitialization~\citep{talbi2009metaheuristics} and weakened convergence pressure as the objective space grows~\citep{fleming2005many}.

Neural combinatorial optimization methods have emerged as a scalable 
alternative \citep{sener2018multi, lin2019pareto}, leveraging learned policies or heuristics to produce high-quality solutions at inference time~\citep{bengio2021machine, veseli2022learning, 
kool2018attention}. Autoregressive approaches~\citep{vinyals2015pointer, 
nazari2018reinforcement} and non-autoregressive methods~\citep{joshi2019efficient, sun2023difusco} have demonstrated strong performance on static combinatorial problems such as the Traveling Salesman Problem (TSP) and Maximum Independent Set (MIS), but assume fixed objective functions and static constraint structures, limiting their applicability to DMOPs. For example, CDM-PSL~\citep{li2025expensive} learns Pareto set distributions via unconditional generation for expensive Bayesian optimization. DM-DMOEA~\citep{wang2025new} employs diffusion-based population reinitialization under severe environmental shifts. Dl-LSDMOEA~\citep{li2026diffusion} treats evolutionary trajectories as supervised training data, aligning the denoising process with population dynamics to generate optimization paths toward Pareto-optimal solutions in new environments. Despite these advances, existing methods predominantly treat environmental dynamics as perturbations to objective parameters. None address the substantially more challenging class of DMOPs in which the combinatorial problem itself evolves over time, such as changes in job configurations, machine capabilities, precedence relationships, routing flexibility, and resource connectivity from the integration of autonomous assets.

Moreover, since even evaluating a single feasible solution requires resolving an NP-hard subproblem~\citep{lenstra1977complexity, garey1976complexity, brandimarte1993routing}, exhaustive enumeration of the Pareto front scales super-polynomially with problem size, and the number of non-dominated solutions itself can grow exponentially with the number of objectives~\citep{ehrgott2005multicriteria, lust2012multiobjective}. Beyond computational intractability, presenting a full Pareto front to a human decision-maker imposes a significant cognitive burden~\citep{miller1956magical, lindauer2022smac3}, as the number of non-dominated solutions far exceeds the capacity of human comparative judgment. In practice, humans typically possess concrete objective targets, such as a desired makespan or a minimum resilience threshold, rather than preferences over an abstract front~\citep{wierzbicki1980use}. This motivates a fundamentally different formulation: rather than approximating the Pareto front and deferring selection to the human, \textit{can we directly learn the distribution of target decisions given a human-specified objectives}?


In this work, we answer this research question by proposing \textsc{GOAL}, a graph-based objective-aligned diffusion solver that generates decisions over heterogeneous problem instances and constraint ontology, conditioned on human-specified objectives without explicitly constructing the Pareto front. We introduce a relational graph neural network (RGNN), allowing constraint information to propagate selectively according to the nature of each problem instance. We evaluate \textsc{GOAL} on three scheduling benchmarks of increasing constraint complexity, including \textsc{FSP}, \textsc{JSP}, and \textsc{FJSP}, and find that \textsc{GOAL} generalizes across all three problem types without retraining, achieving $100\%$ feasibility and MAPEs below $1.6\%$ on \textsc{JSP} and \textsc{FSP}. Finally, we design a held-out generalization experiment on \textsc{JSP} in which \textsc{GOAL} is evaluated on unseen machine counts not encountered during training, finding that solution quality remains within range of seen configurations with all MAPEs below $0.5\%$, demonstrating that the heterogeneous graph representation captures transferable structural information across dynamic constraints.





\section{Related Work}
\label{gen_inst}

\subsection{Autoregressive solvers}

Recent work has explored large language models (LLMs) for combinatorial scheduling from multiple angles~\citep{yang2023foundation}. Abgaryan et al.~\citep{abgaryan2024llms} introduce the supervised dataset of $120$k instances for \textsc{JSP} and demonstrate that a fine-tuned Phi-3 model achieves competitive optimality gaps. Yu et al.~\citep{yu2026automated} develop \textsc{LSH}, which combines an LLM heuristic generator, a performance evaluator, and an evolutionary search module to automatically construct high-quality dispatching heuristics for \textsc{FSP} and \textsc{JSP} without human intervention. Beyond job shop settings, \c{C}etinkaya et al.~\citep{ccetinkaya2025discovering} leverage LLMs to discover novel dispatch rules for the single-machine total tardiness problem, and Wang et al.~\citep{wang2025multi} propose \textsc{MAEF}, a multi-agent framework in which specialized LLM agents collaborate in a feedback-driven evolutionary loop covering problem definition, solution generation, and result evaluation. Despite these advances, two fundamental limitations constrain the applicability of LLM-based approaches to DMOPs: fine-tuning remains computationally resource-intensive and data-hungry~\citep{abgaryan2024llms, brown2020language, touvron2023llama}, and autoregressive generation incurs quadratic self-attention overhead~\citep{vaswani2017attention} that scales poorly with the number of decisions~\citep{sun2023difusco, li2025generative, joshi2019efficient}, rendering these methods impractical for large-scale or time-critical scheduling instances where low-latency inference is required~\citep{hottung2022efficient, kool2018attention}.

\subsection{Diffusion solvers}

Diffusion models have emerged as a powerful paradigm for combinatorial and continuous optimization ~\citep{ho2020denoising, austin2021structured, song2021scorebased}. DDOM learns conditional generative models from offline data via classifier-free guidance, achieving 10-20\% higher objective values with fewer model evaluations \citep{krishnamoorthy2023diffusion}, while DIFFSG and reward-directed diffusion frameworks extend this to network optimization and unlabeled instance generation under target reward conditions \citep{liang2024diffsg, li2024diffusion}. Diffusion-BBO refines low-quality initializations through online conditional sampling \citep{wu2024diff}. In the combinatorial domain, DIFUSCO casts NP-hard problems as discrete \{0,1\}-vector optimization, substantially narrowing the optimality gap on TSP and MIS benchmarks \citep{sun2023difusco}, with T2T further enabling fast single-step inference through direct noise-to-solution mappings \citep{li2024fast}. LMDM demonstrates that diffusion models can approximate combinatorial optimization in accordance with various constraints, with computational efficiency improving at a logarithmic scale \citep{li2025large}. However, despite their promise, existing approaches are limited in two key respects: they optimize for a single scalar objective rather than conditioning on multiple objectives~\citep{sener2018multi, lin2019pareto}; they assume static problems with no adaptation to dynamic constraint topologies~\citep{paulus2021comboptnet, donti2021dc3}. 

\section{Method}

These gaps collectively motivate \textsc{GOAL}, a unified conditional diffusion solver that generates feasible scheduling decisions over graph-encoded problem instances, conditioned on dynamic constraint topologies and human-specified objective targets, without retraining across problem variants or constraint regimes.

\subsection{Problem definition}
\label{sec:problem}
We consider multi-objective DMOPs and define the decision variable as a binary vector ~\citep{sun2023difusco, paulus2021comboptnet}.
\begin{equation}
\mathbf{x} \in \mathcal{X}_s := \{0,1\}^N,
\end{equation}
where $s$ denotes a problem instance and $N = K^2$ is the dimensionality of the decision space, with $K$ denoting the number of nodes (e.g., job operations in JSP). $\mathbf{x} \in \{0,1\}^N$ encodes the selection of pairwise precedence relations between operations~\citep{joshi2019efficient, zhang2020learning}. Let $\{o_s^i(\mathbf{x})\}_{i=1}^I$ denote $I$ objective functions evaluating the performance of a decision $\mathbf{x}$ (e.g., makespan or resilience in JSSP), and let $\{c_s^j(\mathbf{x})\}_{j=1}^J$ denote $J$ constraint functions. These constraints are typically linear or can be reformulated into linear form. The DMOPs problem is formulated as:


\begin{subequations}
\begin{align}
\min_{\mathbf{x} \in \mathcal{X}_s} \quad & \mathbf{o}_s(\mathbf{x}) := \big(o_s^1(\mathbf{x}), \dots, o_s^I(\mathbf{x})\big) \label{eq:obj} \\
\text{s.t.} \quad & c_s^j(\mathbf{x}) \le 0, \quad j = 1, \dots, J. \label{eq:con}
\end{align}
\end{subequations}

\textbf{Challenges of handling dynamic constraints:} Existing neural solvers incorporate feasibility through an additional objective $\sum_{j=1}^{J}\mathrm{valid}(\mathbf{x}, s)$,
where $\mathrm{valid}(\mathbf{x}, s)=0$ if $c_s^j(\mathbf{x}) \le 0$ and $+\infty$ otherwise. 
The learning objective then minimizes
\begin{equation}
\mathbb{E}_{\mathbf{x}\sim p_\theta(\cdot \mid s)}\big[o_s(\mathbf{x})\big]
=
\sum_{\mathbf{x}\in \mathcal{X}_s}
[o_s(\mathbf{x}) + \mathrm{valid}(\mathbf{x},s) ]\, p_\theta(\mathbf{x}\mid s).    
\end{equation}

However, when multiple constraints $\{c_s^j(\mathbf{x}) \le 0\}_{j=1}^J$ are present, collapsing all constraint violations into a single binary feasibility signal discards the geometric structure of the feasible region. This formulation provides neither gradient information nor decomposable feedback about which constraint is violated or by how much, which is a limitation well-documented in constrained combinatorial optimization~\citep{hottung2022efficient, li2021learning}, where constraint-aware learning signals have been shown to be critical for generalization. Consequently, the contribution of individual constraints to the training loss is indistinguishable, preventing the model from adapting to shifts in constraints~\citep{paulus2021comboptnet}. This lack of decomposability is particularly problematic in dynamic settings, where constraint function $\mathbf{c}_s^j(\cdot)$ varies across problem instances and the feasible region changes accordingly.

To address this limitation, we explicitly model the dynamic constraints by constructing a heterogeneous graph representation $\mathcal{G}_s = (\mathcal{V}, \mathcal{E}^1, \dots, \mathcal{E}^J)$, where each edge type $\mathcal{E}^j$ captures the interactions among decisions $\mathbf{x}$ induced by the $j$-th constraint $\mathbf{c}_s^j(\cdot)$. Incorporating this structured representation allows the model to reason explicitly about variations in constraint geometry across instances, providing decomposable structural signals that a scalar feasibility penalty cannot capture~\citep{paulus2021comboptnet, donti2021dc3}.


\textbf{Challenges of handling multiple objectives.}
In multi-objective optimization, a solution $\mathbf{x}^* \in \mathcal{X}_s$ is Pareto optimal if there does not exist another feasible solution $\mathbf{x} \in \mathcal{X}_s$ such that~\citep{deb2002fast, ehrgott2005multicriteria}
\begin{equation}
o_s^i(\mathbf{x}) \le o_s^i(\mathbf{x}^*), \quad \forall i \in \{1,\dots,I\},
\end{equation}
with strict inequality holding for at least one objective. The set of all such solutions defines the Pareto front $\mathcal{P}_s \subset \mathcal{X}_s$~\citep{fleming2005many}. In practice, selecting a final decision requires additional preference information, typically expressed via a utility function $u(\mathbf{o})$ or a target preference vector $\mathbf{u} \in \mathbb{R}^I$. A standard post-hoc selection strategy first approximates $\mathcal{P}_s$ and then computes:
\begin{equation}
\mathbf{x}^* = \arg\min_{\mathbf{x} \in \mathcal{P}_s} \; \| \mathbf{o}_s(\mathbf{x}) - \mathbf{u} \| \quad \text{or} \quad \mathbf{x}^* = \arg\min_{\mathbf{x} \in \mathcal{P}_s} u(\mathbf{o}_s(\mathbf{x})).
\end{equation}

However, constructing $\mathcal{P}_s$ requires exhaustive exploration of the feasible space $\mathcal{X}_s$, which is computationally intensive for large-scale combinatorial problems. Instead, we consider learning a conditional generative model.
\begin{equation}
p_\theta(\mathbf{x} \mid s, \mathbf{u}, \mathcal{G}_s),
\end{equation}
which directly samples feasible solutions conditioned on both the instance $s$ and the preference vector $\mathbf{u}$~\citep{sener2018multi, navon2021learning}. The objective is to generate solutions satisfying feasibility constraints while aligning with the desired preference:
\begin{equation}
\mathbf{o}_s(\mathbf{x}) \approx \mathbf{u}, \quad \text{subject to } \mathbf{x} \in \mathcal{X}_s^{\mathrm{feasible}},
\end{equation}
without explicitly constructing the Pareto front $\mathcal{P}_s$~\citep{lin2019pareto, navon2021learning}.

\subsection{Diffusion models in \textsc{GOAL}}

We instantiate \textsc{Goal} as a diffusion solver~\citep{ho2020denoising, austin2021structured} that learns the distribution $p_\theta(\mathbf{x}_0 \mid s, \mathbf{u}, \mathcal{G}_s)$ over feasible scheduling decisions, conditioned on the problem instance $s$, the preference vector $\mathbf{u}$, and the heterogeneous constraint graph $\mathcal{G}_s$. This formulation naturally addresses both challenges identified in Section~\ref{sec:problem}: the graph-conditioned reverse process incorporates decomposable constraint signals at every denoising step, while the conditional generative objective directly synthesizes preference-aligned decisions without enumerating the Pareto front~\citep{ho2022classifier, sun2023difusco}.

The generative process is defined by a reverse Markov chain:
\begin{equation}
p_\theta(\mathbf{x}_{0:T} \mid s, \mathbf{u}, \mathcal{G}_s) = p(\mathbf{x}_T)\prod_{t=1}^{T} p_\theta(\mathbf{x}_{t-1} \mid \mathbf{x}_t, s, \mathbf{u}, \mathcal{G}_s),
\end{equation}
which gradually denoises $\mathbf{x}_T \sim p(\mathbf{x}_T)$ toward a qualified decision $\mathbf{x}_0$ under the conditioning signal $(s, \mathbf{u}, \mathcal{G}_s)$. The corresponding forward process
\begin{equation}
q(\mathbf{x}_{1:T} \mid \mathbf{x}_0) = \prod_{t=1}^{T} q(\mathbf{x}_t \mid \mathbf{x}_{t-1})
\end{equation}
progressively corrupts $\mathbf{x}_0$ into noise. Training minimizes the variational upper bound on the negative log-likelihood~\citep{ho2020denoising}:
\begin{align}
\mathbb{E}\big[-\log p_\theta(\mathbf{x}_0 \mid s, \mathbf{u}, \mathcal{G}_s)\big]
&\leq\;
\mathbb{E}_q \Big[
\sum_{t=2}^{T}
D_{\mathrm{KL}}\big(
q(\mathbf{x}_{t-1} \mid \mathbf{x}_t, \mathbf{x}_0)
\;\|\;
p_\theta(\mathbf{x}_{t-1} \mid \mathbf{x}_t, s, \mathbf{u}, \mathcal{G}_s)
\big) \notag \\
&\quad - \log p_\theta(\mathbf{x}_0 \mid \mathbf{x}_1, s, \mathbf{u}, \mathcal{G}_s)
\Big] + C.
\end{align}
where $C$ is a constant independent of $\theta$.

\paragraph{Discrete diffusion}
Since scheduling decisions are naturally represented as binary precedence relations, we adopt a discrete denoising diffusion process~\citep{austin2021structured, sun2023difusco} over $\mathbf{x}_0 \in \{0,1\}^{N}$. The forward process independently corrupts each edge toward a maximally uninformative $\operatorname{Bern}(0.5)$ prior:
\begin{equation}
  q\!\left(\mathbf{x}_t \mid \mathbf{x}_0\right)
  = \prod_{i,j}
    \operatorname{Bern}\!\left(
      x_t^{ij};\;
      \bar{\alpha}_t\, x_0^{ij} + \tfrac{1-\bar{\alpha}_t}{2}
    \right),
  \label{eq:forward}
\end{equation}
where $\bar{\alpha}_t = \prod_{s=1}^{t}(1-\beta_s)$ and $\beta_t$ denotes the 
noise schedule.

\paragraph{Training objective}
The denoising network is trained to generate the decision $\hat{\mathbf{x}}_0$ directly, with the reverse 
transition obtained by marginalizing over $\hat{\mathbf{x}}_0$:
\begin{equation}
p_\theta(\mathbf{x}_{t-1} \mid \mathbf{x}_t, s, \mathbf{u}, \mathcal{G}_s)
=
\sum_{\hat{\mathbf{x}}_0}
q(\mathbf{x}_{t-1} \mid \mathbf{x}_t, \hat{\mathbf{x}}_0)\,
p_\theta(\hat{\mathbf{x}}_0 \mid \mathbf{x}_t, s, \mathbf{u}, \mathcal{G}_s).
\end{equation}
The network is optimized via a binary cross-entropy objective over edges:
\begin{equation}
  \mathcal{L}
  = \mathbb{E}_{t \sim \mathcal{U}[1,T],\;
                \mathbf{x}_0,\;
                \mathbf{x}_t \sim q(\mathbf{x}_t \mid \mathbf{x}_0)}\!\left[
      \operatorname{BCE}\!\left(
        p_\theta(\hat{\mathbf{x}}_0 \mid \mathbf{x}_t, s, \mathbf{u}, \mathcal{G}_s),\;
        \mathbf{x}_0
      \right)
    \right],
  \label{eq:loss}
\end{equation}
where $\operatorname{BCE}$ denotes binary cross-entropy with logits and $p_\theta(\hat{\mathbf{x}}_0 \mid \mathbf{x}_t, s, \mathbf{u}, \mathcal{G}_s)$ is implemented by the customized neural network described in Section~\ref{sec:model-structure}. Gradients are clipped to unit $\ell_2$ norm for stable training.




\subsection{Relational graph-based conditional denoising network}
\label{sec:model-structure}

The denoising network $p_\theta$ parameterizes the conditional distribution of the qualified decisions $\hat{\mathbf{x}}_0 \in \{0,1\}^{N}$ given the noisy observation $\mathbf{x}_t$, the diffusion timestep $t$, the problem instance $s$, the constraint graph $\mathcal{G}_s$, and the objective vector $\mathbf{u}$:
$\hat{\mathbf{x}}_0 = p_\theta(\mathbf{x}_t, t, s, \mathbf{u}, \mathcal{G}_s)$. To effectively incorporate heterogeneous conditioning signals, we design a hybrid architecture that processes different input modalities using specialized encoders. For numerical and global features associated with the problem instance and objective conditioning, we choose a multilayer perceptron (MLP) encoder~\citep{hornik1989multilayer}. In contrast, to explicitly model structural dependencies induced by constraints, we represent $\mathcal{G}_s$ as a relational graph and process it using a RGNN~\citep{schlichtkrull2018modeling, battaglia2018relational}, in which each edge type corresponds to a distinct constraint type and messages are propagated selectively per relation type~\citep{hamilton2017inductive}.



\paragraph{Conditioning streams}
Three independent encoders produce conditioning signals injected into every GNN layer~\citep{dhariwal2021diffusion, perez2018film}. The diffusion timestep $t$ is embedded using a sinusoidal positional encoding~\citep{vaswani2017attention}, followed by a two-layer MLP to obtain $\mathbf{e}_t \in \mathbb{R}^{d}$. The problem instance $s$ (e.g., processing times and machine assignments in \textsc{JSP}) is flattened and processed by a three-layer MLP with SiLU activations~\citep{hendrycks2016gaussian}, yielding $\mathbf{e}_s \in \mathbb{R}^{d}$. The target objective vector $\mathbf{u} \in \mathbb{R}^I$ is encoded via a two-layer MLP to produce $\mathbf{e}_o \in \mathbb{R}^{d}$. All three embeddings are broadcast and injected into every GNN layer via affine feature modulation~\citep{perez2018film}, allowing the denoising trajectory to be jointly steered by the problem instance and human-specified objectives.


\paragraph{Node and edge initialization}
Node features are initialized from the degree statistics of $\mathcal{G}_s$. For each node $k$, the degree counts are concatenated and projected to $\mathbb{R}^{H}$ via a learnable linear map~\citep{hamilton2017inductive}. Edge features for each directed edge $(k,k')$ are initialized by stacking the noisy adjacency value $x_t^{kk'} \in \mathbb{R}$ with binary structural indicators of $\mathcal{G}_s$, yielding a multi-dimensional feature vector that is projected to $\mathbb{R}^{H}$.

\paragraph{RGNN layers}
We model the constraint structure as a heterogeneous relational graph $\mathcal{G}_s = (\mathcal{V}, \{\mathcal{E}^j\}_{j \in J})$, where each edge type $j \in J$ corresponds to a distinct constraint type~\cite{schlichtkrull2018modeling}. This design allows the same pair of operations $(k,k')$ to interact differently depending on the constraint type to which their interaction belongs. For instance, in \textsc{JSP}, job-precedence constraints and machine-conflict constraints between the same pair of operations are encoded as two separate edge types and propagated through relation-specific parameters~\cite{schlichtkrull2018modeling}. The message-passing scheme of RGNN follows the general graph network of~\cite{gilmer2017neural, battaglia2018relational}, extended to dynamic constraints.

\paragraph{Anisotropic node aggregation}
At layer $\ell$, node embeddings $\mathbf{h}_k^\ell \in \mathbb{R}^H$ and constraint-specific edge $(k,k')$ embeddings $\mathbf{e}_{kk'}^{j,\ell} \in \mathbb{R}^H$ are updated as follows. A sigmoid gate $\boldsymbol{\sigma}_{kk'}^{j} = \sigma(\mathbf{e}_{kk'}^{j,\ell}) \in (0,1)^H$ controls message strength~\citep{li2016gated}. Messages are aggregated separately over each constraint neighborhood $\mathcal{N}_j(k)$, $j \in \{1,\dots,J\}$:
\begin{equation}
  \mathbf{a}_k^{j}
    = \sum_{k' \in \mathcal{N}_j(k)}
        \boldsymbol{\sigma}_{kk'}^{j}
        \odot \mathbf{V}_j\,\mathbf{h}_{k'}^\ell,
\end{equation}
where $\mathbf{V}_j \in \mathbb{R}^{H \times H}$ is learnable and $\odot$ denotes the Hadamard product. The node representation is updated as:
\begin{equation}
  \mathbf{h}_k^{\ell+1}
    = \mathbf{h}_k^{\ell}
    + \operatorname{ReLU}\!\Bigl(
        \operatorname{BN}\!\Bigl(
          \mathbf{U}^\ell\mathbf{h}_k^{\ell}
          + \mathbf{W}_m
            \textstyle\Big\|_{j=1}^{J}\,\mathbf{a}_k^{j}
          + \mathbf{W}_t\mathbf{e}_t
          + \mathbf{W}_s\mathbf{e}_s
          + \mathbf{W}_o\mathbf{e}_o
        \Bigr)
      \Bigr).
  \label{eq:node_update}
\end{equation}

\paragraph{Edge update.}
Edge features under constraint type $j$ are updated as:
\begin{equation}
\tilde{\mathbf{e}}_{kk'}^{j,\ell+1}
=
\mathbf{P}_j^\ell\,\mathbf{e}_{kk'}^{j,\ell}
+
\mathbf{Q}_j^\ell\,\mathbf{h}_k^\ell
+
\mathbf{R}_j^\ell\,\mathbf{h}_{k'}^\ell,
\end{equation}
\begin{equation}
\mathbf{e}_{kk'}^{j,\ell+1}
=
\mathbf{e}_{kk'}^{j,\ell}
+
\operatorname{MLP}_e\!\Bigl(
\operatorname{BN}\!\bigl(\tilde{\mathbf{e}}_{kk'}^{j,\ell+1}\bigr)
+
\mathbf{W}_t\,\mathbf{e}_t
+
\mathbf{W}_s\,\mathbf{e}_s
+
\mathbf{W}_o\,\mathbf{e}_o
\Bigr),
\end{equation}
where $\mathbf{P}_j^\ell, \mathbf{Q}_j^\ell, \mathbf{R}_j^\ell \in 
\mathbb{R}^{H \times H}$ are learnable constraint-specific parameters. After 
$L$ layers, a shared edge readout 
($\operatorname{BN} \to \operatorname{ReLU} \to \operatorname{Linear}(H,1)$) 
produces a scalar denoising logit for each edge $(k,k')$:
\begin{equation}
\hat{x}_{kk'}^{0}
=
\operatorname{MLP}_{\mathrm{out}}\!\Bigl(
\textstyle\Big\|_{j=1}^{J}\,\mathbf{e}_{kk'}^{j,L}
\Bigr).
\label{eq:readout}
\end{equation}

\subsection{Classifier-free guidance}
To enable controllable generation with respect to the objective $\mathbf{u}$, we adopt classifier-free guidance (CFG)~\citep{ho2022classifier}. During training, the conditioning vector is stochastically dropped with probability $p_{\mathrm{drop}}$ by replacing $\mathbf{u}$ with $\mathbf{0}$, which trains the denoising network $p_\theta(\mathbf{x}_t, t, \mathbf{s}, \mathbf{G}, \mathbf{u})$ to perform both conditional and unconditional denoising within a single set of parameters.

At inference time, we obtain two predictions from the shared model: a conditional estimate $\hat{\mathbf{x}}_0^{\mathrm{cond}} = p_\theta(\mathbf{x}_t, t, \mathbf{s}, \mathbf{G}, \mathbf{u})$ and an unconditional estimate $\hat{\mathbf{x}}_0^{\mathrm{uncond}} = p_\theta(\mathbf{x}_t, t, \mathbf{s}, \mathbf{G}, \mathbf{0})$. The guided prediction is then formed as
\begin{equation}
\hat{\mathbf{x}}_0^{\mathrm{guided}}
=
\hat{\mathbf{x}}_0^{\mathrm{uncond}}
+
\gamma \bigl(
\hat{\mathbf{x}}_0^{\mathrm{cond}}
-
\hat{\mathbf{x}}_0^{\mathrm{uncond}}
\bigr),
\end{equation}
where $\gamma \geq 1$ denotes the guidance strength. $p_\theta$ outputs logits for each edge, and the guided logits are passed through a sigmoid to obtain probabilities, which are then used to parameterize the reverse transition $q(\mathbf{x}_{t-1} \mid \mathbf{x}_t, \hat{\mathbf{x}}_0^{\mathrm{guided}})$. At the final step ($t=0$), the binary solution is obtained by thresholding the predicted probabilities.

\subsection{Denoising schedule for fast inference} 
\label{sec:inference}
To accelerate inference, we adopt a discrete variant of denoising diffusion implicit models (DDIM)~\citep{song2021denoising}, which reduces the number of reverse diffusion steps by operating on a sub-sequence of timesteps~\citep{watson2022learning}. Formally, instead of performing denoising over the full Markov chain $\{1,\dots,T\}$, we construct an increasing 
subsequence $\tau = \{\tau_1, \dots, \tau_M\}$, with $\tau_1 = 1$, $\tau_M = T$, 
and $M \ll T$. The reverse process is then defined over the reduced trajectory $\mathbf{x}_{\tau_M} \rightarrow \cdots \rightarrow \mathbf{x}_{\tau_1}$, where each transition directly models $p_\theta(\mathbf{x}_{\tau_{i-1}} \mid \mathbf{x}_{\tau_i}, s, \mathbf{u}, \mathcal{G}_s)$, thereby reducing intermediate denoising steps. We consider both linear and cosine schedules for constructing $\tau$. The linear schedule uniformly subsamples timesteps via $\tau_i = \lfloor c\, i \rfloor$, while the cosine schedule allocates more steps in the low-noise regime~\citep{nichol2021improved}, defined as
\begin{equation}
\tau_i = \left\lfloor \cos\!\left(\frac{(1 - c_i)\pi}{2}\right)\cdot T \right\rfloor,
\quad c_i \in [0,1].
\end{equation}

This allocation is motivated by the observation that score-based models exhibit greater sensitivity to denoising updates at low noise levels~\citep{song2021denoising}, so concentrating evaluations yields improved sample quality for a fixed inference budget.

\subsection{Decoding strategies}


At inference time, we decode the predicted binary decision matrix $\mathbf{x} \in \{0,1\}^{N \times N}$, where $N$ is the total number of operations, into a schedule by processing operations in topological order and assigning each operation its earliest feasible start time. For operation $(j,k)$, denoting the $k$-th operation of job $j$ with assigned machine $m_{jk}$ and processing time $p_{jk} > 0$, the start time is given by
\begin{equation}
    s_{jk} = \max\!\bigl(s_{j,k-1} + p_{j,k-1},\; r_{m_{jk}}\bigr),
\end{equation}
where $r_{m_{jk}} \in \mathbb{R}_{\geq 0}$ denotes the earliest time at which machine $m_{jk}$ becomes available. We set $s_{j,0} = 0$ and $p_{j,0} = 0$ by convention for the dummy predecessor of each job's first operation. After scheduling operation $(j,k)$, we update the machine availability as $r_{m_{jk}} \leftarrow s_{jk} + p_{jk}$.

\section{Experiments}
\label{sec:experiments}
\paragraph{Datasets}
We generate and label training data using synthetic JSP, FSP, and FJSP instances, parameterized by the number of jobs $n_j$, operations per job $n_o$, and machines $n_m$. Across all problem variants, we fix $n_o = 3$ and vary $n_j \in \{5, \dots, 20\}$ and $n_m \in \{3, \dots, 10\}$, with processing times independently sampled from a uniform distribution $[1,5]$. The total number of operations is given by $K = n_j \times n_o$.  Each instance is solved using the OR-Tools CP-SAT solver, with $200$ feasible schedules enumerated via a solution callback. The data stores both the raw problem specifications and the corresponding feasible schedules for downstream graph construction. Each feasible schedule is represented as a directed graph encoding precedence and resource constraints. 


\paragraph{Model settings} The denoising network is parameterized as a RGNN with hidden dimension $H = 128$, embedding dimension $d_e = 256$, and conditioning dimension $d_c = 256$. The network consists of $L = 12$ message-passing layers operating on node features $\mathbf{h}_i^\ell \in \mathbb{R}^{H}$ and edge features $\mathbf{e}_{ij}^\ell \in \mathbb{R}^{H}$. 
We train the model using a Bernoulli diffusion process with horizon $T = 1{,}000$ with a linear schedule with $\beta_1 = 10^{-4}$ and $\beta_T = 0.02$. Optimization is performed using AdamW with learning rate $10^{-4}$ and weight decay $10^{-4}$ for $25$ epochs with batch size $64$, and gradients are clipped to an $\ell_2$ norm of $1.0$ for stability. We employ classifier-free guidance by dropping the conditioning vector with probability $p_{\mathrm{drop}} = 0.1$ during training and use a guidance scale $\gamma = 2.0$ at inference to control the strength of conditioning.

\paragraph{Training/Test split}
The training corpus comprises 5{,}000 problem instances per problem type, each yielding 200 solver-generated decisions with corresponding objective targets $(C_{\max}, R)$. For evaluation, 100 problem instances are uniformly sampled at random, with target objectives drawn independently to ensure that the queried combinations are largely unseen during training, providing a controller assessment of out-of-distribution generalization. 

 

\paragraph{Evaluation metrics}
We evaluate decision performance along two objectives: \textit{makespan} $C_{\max}$, defined as the total completion time of all jobs, and \textit{resilience} $R$, defined as the mean temporal slack across all operations, $R \;=\; \frac{\sum_{k \in K} \left(\mathrm{LS}_k - \mathrm{ES}_k\right)}{C_{\max}},$ where $\mathrm{ES}_k$ and $\mathrm{LS}_k$ denote the earliest and latest feasible start times of operation $k$, respectively. For each method, we report the \textit{optimality gap}, measured as the mean absolute percentage error (MAPE) between the best solution found and the objective target for both $C_{\max}$ and $R$. Solution diversity is quantified via the \textit{duplication rate}, defined as the fraction of generated decision candidates that are near-identical, reflecting the effective coverage of the solution space. Feasibility rate denotes the proportion of generated decisions that satisfy the dynamic constraints. Inference efficiency is measured by \textit{time-to-$\epsilon$}, the wall-clock time required to produce the first solution satisfying $|C_{\max} - C_{\max}^{*}| / C_{\max}^{*} \leq \epsilon$ and $|R - R^{*}| / R^{*} \leq \epsilon$ simultaneously, evaluated at $\epsilon \in \{5\%, 10\%\}$. For baseline methods, \textit{time-to-$\epsilon$} is reported as the mean wall-clock time required to obtain the first $\epsilon$-feasible decision. For \textsc{GOAL}, it is computed as the average wall-clock time per generated decision that satisfies the $\epsilon$-feasibility criterion.

\section{Results}
\label{sec:result}

\subsection{Scalability analysis}
Since existing methods are designed to minimize the objectives or approximate the Pareto front (for DMOPs) rather than target a specific objective value, a direct performance comparison requires adapting their evaluation criterion. We therefore reformulate the fitness function for both \textsc{NSGA-II} and \textsc{MOEA/D} as a squared relative error with respect to the target $(C^*_{\max}, R^*)$, penalized by constraint violations, enabling a fair comparison against \textsc{GOAL} on the same time-to-$\varepsilon$ metric. 

        
Table~\ref{tab:scalability_comparison} reports that \textsc{Goal} consistently achieves the fastest qualified decision time at both thresholds, with inference times remaining below $78.1$\,ms at the $10\%$ threshold and below $148.8$\,ms at the $5\%$ threshold across all sizes. By contrast, \textsc{NSGA-II} requires between $159.4$\,ms and $3{,}816.4$\,ms, while \textsc{MOEA/D} ranges from $86.1$\,ms to $1{,}628.9$\,ms before failing entirely on $20{\times}3$. At the $5\%$ threshold on $10{\times}3$, \textsc{Goal} requires only $125.0$\,ms compared to $822.3$\,ms for \textsc{NSGA-II} and $1628.9$\,ms for \textsc{MOEA/D}, representing speedups of $6.5{\times}$ and $8.2{\times}$, respectively. This advantage compounds at scale: at $20{\times}3$, \textsc{Goal} requires $148.8$\,ms versus $3{,}816.4$\,ms for \textsc{NSGA-II} — a $25{\times}$ speedup.

\begin{figure}[t]
    \centering
    \includegraphics[width=\linewidth]{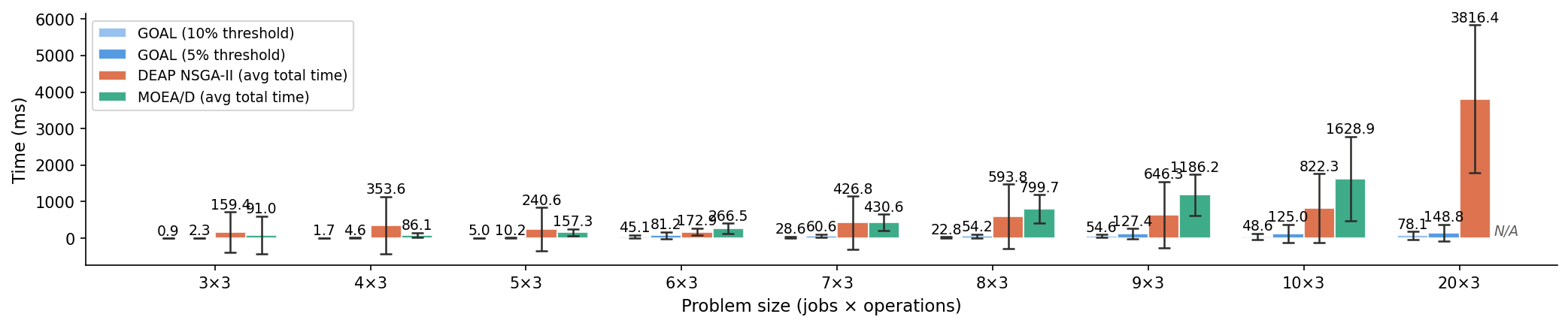}
    \caption{
        Mean time to reach a qualified solution (milliseconds) across JSP instances of job size.
    }
    \label{fig:time_comparison}
\end{figure}

\textsc{Goal} achieves perfect feasibility (100\%) to accommodate the constraints across all nine problem sizes, consistently producing solutions within the 5\% error threshold on both $C_{max}$ and $R$ objectives, while maintaining near-zero MAPE values (below 0.20\% for MS and UR across all sizes), demonstrating strong and stable generalization from small ($3\times3$) to large ($20\times3$) instances. The duplication rate for \textsc{Goal} drops sharply from 45.43\% at $3\times3$ to effectively zero for all instances of size $5\times3$ and above, indicating that the diffusion model generates increasingly diverse solution candidates as problem complexity grows, with the elevated duplication at $3\times3$ attributable to the limited combinatorial search space of that smallest instance.

\begin{table}[t]
\centering
\scriptsize
\setlength{\tabcolsep}{1pt}
\renewcommand{\arraystretch}{1.2}
\caption{Scalability comparison of \textsc{GOAL}, NSGA-II, and MOEA/D across problem sizes. MAPE values are computed exclusively over feasible trials. N/A indicates no feasible solution was found. Results averaged over 100 problem instances.}
\label{tab:scalability_comparison}
\begin{tabular}{llccccccccc}
\toprule
& & \multicolumn{9}{c}{Problem Size} \\
\cmidrule(lr){3-11}
Method & Metric
& $3{\times}3$ & $4{\times}3$ & $5{\times}3$ & $6{\times}3$ & $7{\times}3$
& $8{\times}3$ & $9{\times}3$ & $10{\times}3$ & $20{\times}3$ \\
\midrule
\multirow{3}{*}{NSGA-II}
& $C_{max}$ (\%)
& $0.53_{\pm1.49}$ & $1.19_{\pm1.97}$ & $2.11_{\pm2.08}$ & $2.60_{\pm1.88}$ & $2.11_{\pm1.72}$
& $2.30_{\pm1.54}$ & $2.14_{\pm1.52}$ & $2.25_{\pm1.65}$ & $2.60_{\pm1.48}$ \\
& $R$ (\%)
& $0.55_{\pm1.52}$ & $1.17_{\pm1.95}$ & $2.10_{\pm2.05}$ & $2.58_{\pm1.86}$ & $2.09_{\pm1.70}$
& $2.27_{\pm1.52}$ & $2.12_{\pm1.49}$ & $2.23_{\pm1.63}$ & $2.54_{\pm1.43}$ \\
& Feas. (\%)
& 95.00 & 92.00 & 96.00 & \textbf{100.00} & 97.00
& 95.00 & 97.00 & 97.00 & 96.00 \\
\midrule
\multirow{3}{*}{MOEA/D}
& $C_{max}$ (\%)
& $0.45_{\pm1.36}$ & $1.43_{\pm2.07}$ & $2.12_{\pm2.12}$ & $2.09_{\pm1.94}$ & $2.17_{\pm1.69}$
& $\mathbf{0.00}_{\pm0.00}$ & $\mathbf{0.00}_{\pm0.00}$ & $2.30_{\pm1.59}$ & N/A \\
& $R$ (\%)
& $0.45_{\pm1.36}$ & $1.42_{\pm2.05}$ & $2.09_{\pm2.09}$ & $2.07_{\pm1.91}$ & $2.16_{\pm1.67}$
& $\mathbf{0.00}_{\pm0.00}$ & $\mathbf{0.00}_{\pm0.00}$ & $2.27_{\pm1.56}$ & N/A \\
& Feas. (\%)
& 99.00 & \textbf{100.00} & \textbf{100.00} & \textbf{100.00} & \textbf{100.00}
& 99.00 & \textbf{100.00} & \textbf{100.00} & N/A \\
\midrule
\multirow{4}{*}{GOAL}
& $C_{max}$ (\%)
& $\mathbf{0.00}_{\pm0.00}$ & $\mathbf{0.00}_{\pm0.00}$ & $\mathbf{0.06}_{\pm0.59}$ & $\mathbf{0.19}_{\pm0.85}$ & $\mathbf{0.03}_{\pm0.30}$
& $0.03_{\pm0.29}$ & $0.03_{\pm0.29}$ & $\mathbf{0.16}_{\pm0.95}$ & $\mathbf{0.02}_{\pm0.16}$ \\
& $R$ (\%)
& $\mathbf{0.00}_{\pm0.00}$ & $\mathbf{0.00}_{\pm0.00}$ & $\mathbf{0.06}_{\pm0.56}$ & $\mathbf{0.19}_{\pm0.83}$ & $\mathbf{0.03}_{\pm0.31}$
& $0.03_{\pm0.29}$ & $0.03_{\pm0.28}$ & $\mathbf{0.16}_{\pm0.88}$ & $\mathbf{0.02}_{\pm0.16}$ \\
& Feas. (\%)
& \textbf{100.00} & \textbf{100.00} & \textbf{100.00} & \textbf{100.00} & \textbf{100.00}
& \textbf{100.00} & \textbf{100.00} & \textbf{100.00} & \textbf{100.00} \\
\cmidrule(lr){2-11}
& Dup. (\%)
& $45.43_{\pm26.17}$ & $2.03_{\pm4.41}$ & $0.00_{\pm0.00}$ & $0.00_{\pm0.00}$ & $0.00_{\pm0.00}$
& $0.00_{\pm0.00}$ & $0.00_{\pm0.00}$ & $0.00_{\pm0.00}$ & $0.00_{\pm0.00}$ \\
\bottomrule
\end{tabular}
\end{table}

\subsection{Generalization across problem variants}

To assess generalization across structurally distinct scheduling variants, we evaluate \textsc{GOAL} on three scheduling problem types: \textsc{FSP}, which enforces a fixed global machine ordering shared across all jobs; \textsc{JSP}, which permits flexible sequencing of operations on each machine; and \textsc{FJSP}, which further allows each operation to be assigned to any machine within a predefined eligible subset, thereby inducing richer constraint coupling over a larger feasible space. For each problem type, we sample $3{,}000$ training instances with $\lvert\text{jobs}\rvert \in \{5, 7, 10\}$ and evaluate on $100$ held-out instances per type following the descriptions in Section~\ref{sec:experiments}.

\begin{figure}[h!]
\begin{subfigure}[t]{0.52\textwidth}
\captionsetup{labelformat=empty}
\begingroup
  \renewcommand{\thesubfigure}{}
  \refstepcounter{table}
  \protected\edef\@currentlabel{\thetable}
  \label{tab:results}
\endgroup
\caption{\textbf{Table~\thetable:} Summary of $C_{max}$ MAPE, $R$ MAPE, averaged duplication rate, and feasibility rate across JSP, FSP, FJSP. Mean$_{\pm\text{std}}$ over 100 trials.}
\centering
\scriptsize
\setlength{\tabcolsep}{2pt}
\renewcommand{\arraystretch}{1.2}
\resizebox{\textwidth}{!}{%
\begin{tabular}{llcccc}
\toprule
& & \multicolumn{4}{c}{Metric} \\
\cmidrule(lr){3-6}
Type & Instance & $C_{max}$ (\%) & $R$ (\%) & Dup.\ Rate (\%) & Feas.\ (\%) \\
\midrule
\multirow{3}{*}{JSP}
& $5{\times}3$  & $0.00_{\pm0.00}$ & $0.00_{\pm0.00}$   & $0.00_{\pm0.00}$ & $100.00_{\pm0.00}$ \\
& $7{\times}3$  & $0.15_{\pm0.77}$ & $0.15_{\pm0.74}$   & $0.00_{\pm0.00}$ & $100.00_{\pm0.00}$ \\
& $10{\times}3$ & $1.13_{\pm2.31}$ & $1.09_{\pm2.14}$   & $0.00_{\pm0.00}$ & $100.00_{\pm0.00}$ \\
\midrule
\multirow{3}{*}{FSP}
& $5{\times}3$  & $0.32_{\pm1.11}$ & $0.31_{\pm1.07}$   & $0.00_{\pm0.00}$ & $100.00_{\pm0.00}$ \\
& $7{\times}3$  & $0.00_{\pm0.00}$ & $0.00_{\pm0.00}$   & $0.00_{\pm0.00}$ & $100.00_{\pm0.00}$ \\
& $10{\times}3$ & $1.54_{\pm3.65}$ & $1.41_{\pm3.16}$   & $0.00_{\pm0.00}$ & $100.00_{\pm0.00}$ \\
\midrule
\multirow{3}{*}{FJSP}
& $5{\times}3$  & $3.67_{\pm7.28}$ & $12.53_{\pm10.83}$ & $0.01_{\pm0.06}$ & $100.00_{\pm0.00}$ \\
& $7{\times}3$  & $3.98_{\pm8.99}$ & $10.17_{\pm8.41}$  & $0.00_{\pm0.00}$ & $100.00_{\pm0.00}$ \\
& $10{\times}3$ & $6.97_{\pm9.72}$ & $11.80_{\pm7.86}$  & $0.00_{\pm0.00}$ & $100.00_{\pm0.00}$ \\
\bottomrule
\end{tabular}}
\end{subfigure}
\hfill
\begin{subfigure}[t]{0.48\textwidth}
\captionsetup{labelformat=empty, position=below}
\begingroup
  \renewcommand{\thesubfigure}{}
  \refstepcounter{figure}
  \protected\edef\@currentlabel{\thefigure}
  \label{fig:time_per_problem_type}
\endgroup
\centering
\includegraphics[width=\textwidth]{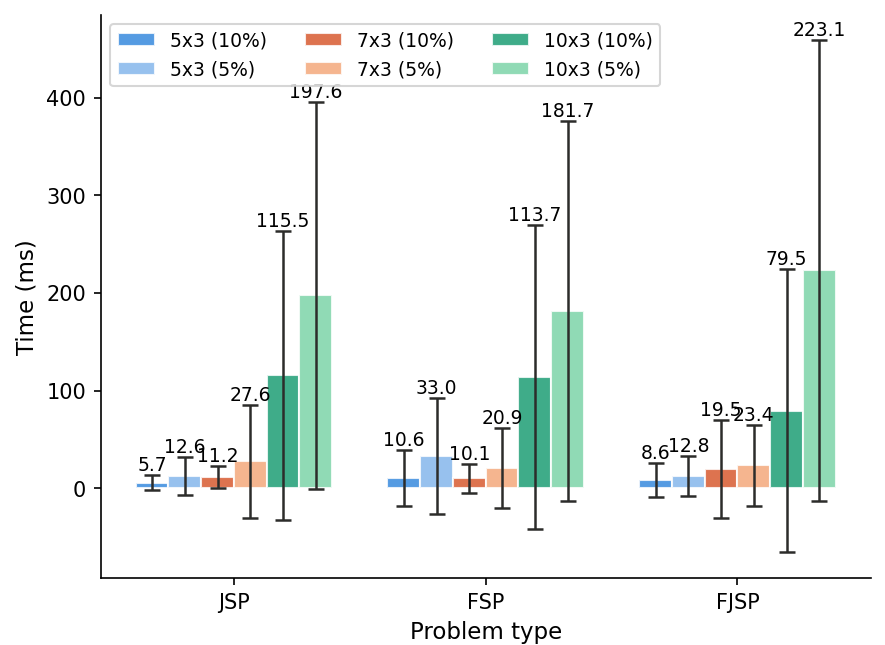}
\caption{\textbf{Figure~\thefigure:} Time-to-$\epsilon$ across problem types and instances. Error bars denote std over 100 instances.}
\end{subfigure}
\end{figure}

Tab.~\ref{tab:results} reports $C_{max}$ MAPE, $R$ MAPE, duplication rate, and feasibility across JSP, FSP, and FJSP for three instance sizes. \textsc{Goal} achieves $100\%$ feasibility with zero duplicate actions across all settings. For JSP and FSP, both MAPEs remain below $1.6\%$ at all scales, while FJSP yields higher errors (up to $6.97\%$ $C_{max}$ MAPE and $12.53\%$ $R$ MAPE). This is consistent with the broader objective span induced by flexible machine assignment: given the same candidate budget, the additional routing degree of freedom in FJSP widens the range of attainable $C_{\max}$ and $R$ values, so the expected distance between the best candidate found and the true optimum grows, directly inflating the reported MAPE. Furthermore, the increased span reduces the density of candidates per unit of objective range, which may impair the model's ability to discriminate between decisions of similar but distinct objective value.

Fig.~\ref{fig:time_per_problem_type} reports time-to-$\varepsilon$ across problem types and instance sizes. Decision times remain comparable across \textsc{JSP}, \textsc{FSP}, and \textsc{FJSP} at each fixed scale, while growing notably from $5{\times}3$ to $10{\times}3$ instances. This pattern holds under both $\varepsilon \in \{5\%, 10\%\}$ thresholds, indicating that inference efficiency is primarily governed by instance scale rather than problem type.


\subsection{Generalization across new constraints}

To evaluate generalization to new constraints, we design an experiment on \textsc{JSP} in which \textsc{GOAL} is trained with an incomplete set of machine counts. Tab.~\ref{tab:results_machines} reports model performance across both provided and held-out machine counts (shaded gray). \textsc{Goal} maintains $100\%$ feasibility across all configurations, and MAPEs on unseen machine counts remain consistently comparable to those on seen counts. For instance, on $10{\times}3$ instances, MS MAPE at unseen $n_m{=}7$ ($0.12_{\pm0.69}$) and $n_m{=}9$ ($0.22_{\pm0.99}$) are within range of seen configurations, with all errors remaining below $0.5\%$. This indicates that \textsc{GOAL} generalizes effectively to constraints not encountered during training, without degradation in solution quality or feasibility. 

\begin{figure}[h!]
\begin{subfigure}[t]{0.64\textwidth}
\captionsetup{labelformat=empty}
\begingroup
  \renewcommand{\thesubfigure}{}
  \refstepcounter{table}
  \protected\edef\@currentlabel{\thetable}
  \label{tab:results_machines}
\endgroup
\caption{\textbf{Table~\thetable:} Summary of model performance across dynamic constraints. Mean$_{\pm\text{std}}$ over 100 trials. Light grey indicates the unseen machine numbers during training}
\vspace{4pt}
\centering
\scriptsize
\setlength{\tabcolsep}{2pt}
\renewcommand{\arraystretch}{1.2}
\resizebox{\textwidth}{!}{%
\begin{tabular}{l>{\columncolor{white}}c>{\columncolor{lightgray}}c>{\columncolor{white}}c
                 >{\columncolor{white}}c>{\columncolor{lightgray}}c>{\columncolor{white}}c>{\columncolor{white}}c}
\toprule
& \multicolumn{3}{c}{$5{\times}3$}
& \multicolumn{4}{c}{$7{\times}3$} \\
\cmidrule(lr){2-4}\cmidrule(lr){5-8}
Metric & $n_m{=}3$ & $n_m{=}4$ & $n_m{=}5$
& $n_m{=}4$ & $n_m{=}5$ & $n_m{=}6$ & $n_m{=}7$ \\
\midrule
$C_{max}$ (\%) & $0.00_{\pm0.00}$ & $0.00_{\pm0.00}$ & $0.00_{\pm0.00}$
& $0.31_{\pm1.08}$ & $0.05_{\pm0.45}$ & $0.00_{\pm0.00}$ & $0.00_{\pm0.00}$ \\
$R$ (\%) & $0.00_{\pm0.00}$ & $0.00_{\pm0.00}$ & $0.00_{\pm0.00}$
& $0.30_{\pm1.04}$ & $0.04_{\pm0.44}$ & $0.00_{\pm0.00}$ & $0.00_{\pm0.00}$ \\
Feas.\ (\%)  & N/A & N/A & N/A
& $100_{\pm0}$ & $100_{\pm0}$ & $100_{\pm0}$ & $100_{\pm0}$ \\
\midrule
& \multicolumn{7}{c}{$10{\times}3$} \\
\cmidrule(lr){2-8}
Metric & $n_m{=}4$ & $n_m{=}5$ & $n_m{=}6$ & \cellcolor{lightgray}$n_m{=}7$ & \cellcolor{white}$n_m{=}8$ & \cellcolor{lightgray}$n_m{=}9$ & $n_m{=}10$ \\
\midrule
$C_{max}$ (\%) & $0.46_{\pm1.28}$ & $0.46_{\pm1.62}$ & $0.08_{\pm0.59}$
& \cellcolor{lightgray}$0.12_{\pm0.69}$ & $\cellcolor{white}0.13_{\pm0.74}$ & \cellcolor{lightgray}$0.22_{\pm0.99}$ & $0.27_{\pm2.22}$ \\
$R$ (\%) & $0.45_{\pm1.23}$ & $0.43_{\pm1.52}$ & $0.08_{\pm0.57}$
& \cellcolor{lightgray}$0.12_{\pm0.68}$ & $\cellcolor{white}0.13_{\pm0.72}$ & \cellcolor{lightgray}$0.22_{\pm0.95}$ & $0.23_{\pm1.85}$ \\
Feas.\ (\%)  & $100_{\pm0}$ & $100_{\pm0}$ & $100_{\pm0}$
& \cellcolor{lightgray}$100_{\pm0}$ & \cellcolor{white}$100_{\pm0}$ & \cellcolor{lightgray}$100_{\pm0}$ & $100_{\pm0}$ \\
\bottomrule
\end{tabular}}
\end{subfigure}
\hfill
\begin{subfigure}[t]{0.35\textwidth}
\captionsetup{labelformat=empty, position=below}
\begingroup
  \renewcommand{\thesubfigure}{}
  \setcounter{figure}{3}
  \refstepcounter{figure}
  \protected\edef\@currentlabel{\thefigure}
  \label{fig:time_per_machines}
\endgroup
\centering
\includegraphics[width=\textwidth]{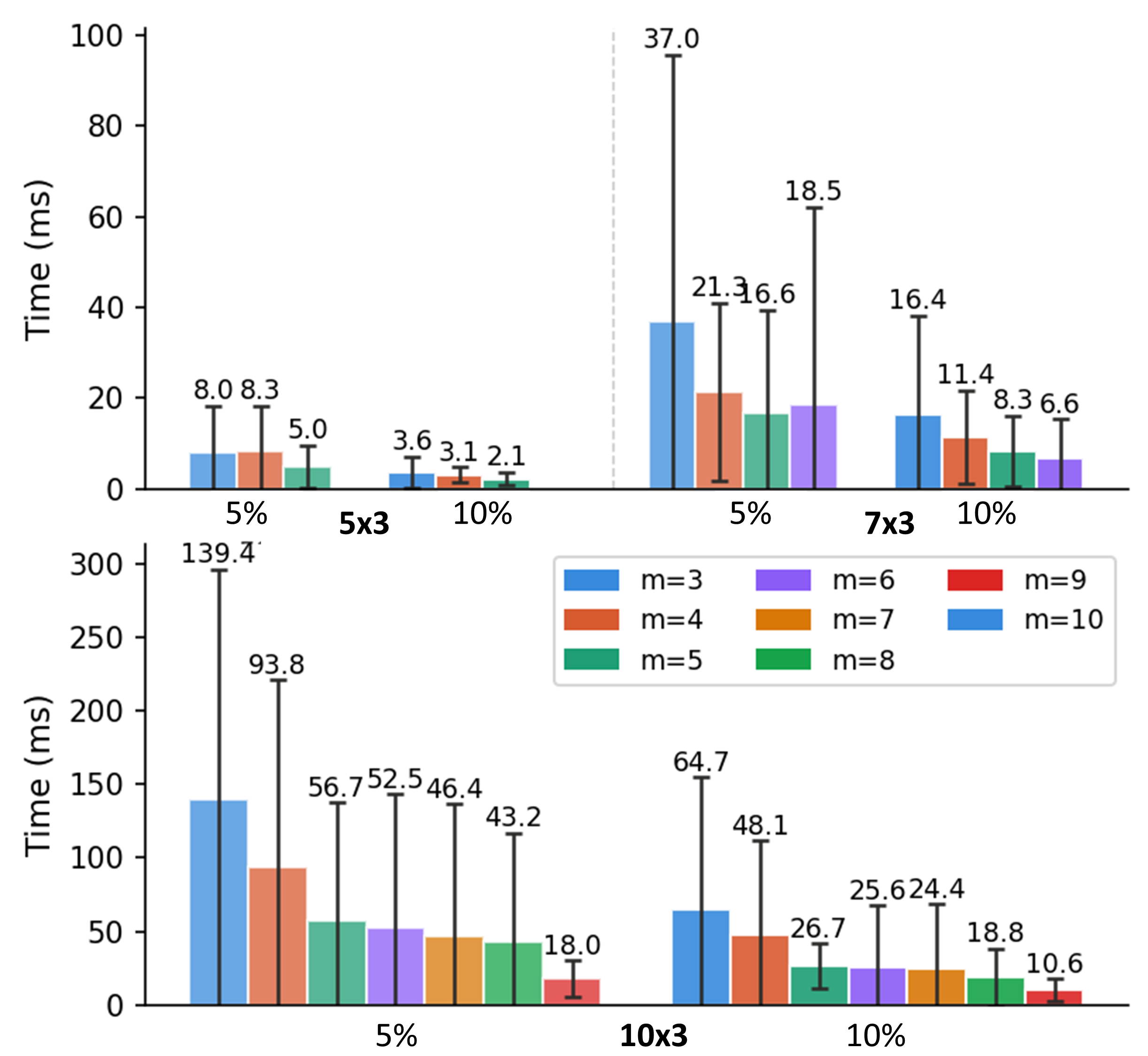}
\caption{\textbf{Figure~\thefigure:} Time to qualified decision grouped by machine counts.}
\end{subfigure}
\end{figure}
Fig.~\ref{fig:time_per_machines} shows that time-to-$\varepsilon$ tends to decrease as the number of machines increases. This is consistent with the structure of \textsc{JSP}: with more machines available relative to the number of jobs, each machine handles fewer operations on average, reducing contention and broadening the set of feasible orderings. The resulting decrease in scheduling flexibility and feasible space allows \textsc{GOAL} to identify a qualified solution more rapidly, suggesting that decision time is sensitive not only to job scale but also to the degree of resource contention induced by the machine-to-job ratio.

\section{Conclusions}

We proposed \textsc{Goal}, a conditional diffusion model over graph representations for multi-objective scheduling under dynamic constraints. We introduced a relational graph neural network with type-specific message passing that encodes structurally distinct constraints, enabling \textsc{Goal} to condition jointly on human-specified objectives and dynamic constraint topologies. \textsc{Goal} achieves $100\%$ feasibility and near-zero MAPE across \textsc{JSP}, \textsc{FSP}, and \textsc{FJSP}, outperforming \textsc{NSGA-II} and \textsc{MOEA/D} by up to $25\times$ in inference speed while generalizing to unseen problem types and constraint configurations without architectural modification. 

For future work, we would like to extend \textsc{Goal} to a broader class of DMOPs. We would like to incorporate machine nodes explicitly into the heterogeneous graph representation, enabling richer constraint encoding and improving generalizability across problem variants with varying routing flexibility. Additionally, we are interested in investigating auxiliary objective-alignment losses, data augmentation strategies, and adaptive guidance schedules to improve solution accuracy under high-flexibility settings such as \textsc{FJSP}, where the broader decision space currently yields larger MAPEs. Finally, we would like to explore principled graph sparsification strategies to improve scalability to very large-scale instances involving thousands of operations without sacrificing constraint satisfaction.

\section*{Acknowledgements}
This work was supported in part by the National Science Foundation under grants NSF-2531898.












\bibliographystyle{unsrt}
\bibliography{reference}

@article{jiang2022evolutionary,
  title={Evolutionary dynamic multi-objective optimisation: A survey},
  author={Jiang, Shouyong and Zou, Juan and Yang, Shengxiang and Yao, Xin},
  journal={ACM Computing Surveys},
  volume={55},
  number={4},
  pages={1--47},
  year={2022},
  publisher={ACM New York, NY}
}

@article{sun2023difusco,
  title={Difusco: Graph-based diffusion solvers for combinatorial optimization},
  author={Sun, Zhiqing and Yang, Yiming},
  journal={Advances in neural information processing systems},
  volume={36},
  pages={3706--3731},
  year={2023}
}

@inproceedings{ha2023embodied,
  title     = {Embodied Intelligence via Learning and Evolution},
  author    = {Ha, David and Tang, Yujin},
  booktitle = {Advances in Neural Information Processing Systems},
  year      = {2023}
}

@article{lasi2014industry,
  title   = {Industry 4.0},
  author  = {Lasi, Heiner and Fettke, Peter and Kemper, Hans-Georg and
             Feld, Thomas and Hoffmann, Michael},
  journal = {Business and Information Systems Engineering},
  volume  = {6},
  number  = {4},
  pages   = {239--242},
  year    = {2014}
}

@inproceedings{yu2022surprising,
  title     = {The Surprising Effectiveness of {PPO} in Cooperative
               Multi-Agent Games},
  author    = {Yu, Chao and Velu, Akash and Vinitsky, Eugene and Gao,
               Jiaxuan and Wang, Yu and Bayen, Alexandre and Wu, Yi},
  booktitle = {Advances in Neural Information Processing Systems},
  year      = {2022}
}

@inproceedings{zhang2020learning,
  title     = {Learning to Dispatch for Job Shop Scheduling via Deep
               Reinforcement Learning},
  author    = {Zhang, Cong and Song, Wen and Cao, Zhiguang and Zhang,
               Jie and Tan, Puay Siew and Xu, Chi},
  booktitle = {Advances in Neural Information Processing Systems},
  year      = {2020}
}

@inproceedings{liu2023dynamic,
  title     = {Dynamic Task Allocation for Multi-Robot Systems Using
               Graph Neural Networks},
  author    = {Liu, Shengchao and others},
  booktitle = {Advances in Neural Information Processing Systems},
  year      = {2023}
}

@inproceedings{park2021schedulenet,
  title     = {{ScheduleNet}: Learn to Solve Multi-Agent Scheduling Problems
               with Reinforcement Learning},
  author    = {Park, Junyoung and Bakhtiyar, Sanjar and Park, Jinkyoo},
  booktitle = {arXiv preprint arXiv:2106.03051},
  year      = {2021}
}

@article{gronauer2022multi,
  title   = {Multi-Agent Deep Reinforcement Learning: A Survey},
  author  = {Gronauer, Sven and Diepold, Klaus},
  journal = {Artificial Intelligence Review},
  volume  = {55},
  pages   = {895--943},
  year    = {2022}
}

@article{brettel2014virtualization,
  title   = {How Virtualization, Decentralization and Network Building Change
             the Manufacturing Landscape},
  author  = {Brettel, Malte and Friederichsen, Niklas and Keller, Michael and
             Rosenberg, Marius},
  journal = {International Journal of Mechanical, Aerospace, Industrial and
             Mechatronics Engineering},
  volume  = {8},
  number  = {1},
  pages   = {37--44},
  year    = {2014}
}

@inproceedings{reed2022generalist,
  title     = {A Generalist Agent},
  author    = {Reed, Scott and others},
  booktitle = {Transactions on Machine Learning Research},
  year      = {2022}
}

@inproceedings{kaufmann2023champion,
  title     = {Champion-Level Drone Racing Using Deep Reinforcement Learning},
  author    = {Kaufmann, Elia and others},
  journal   = {Nature},
  year      = {2023}
}

@inproceedings{rudin2022learning,
  title     = {Learning to Walk in Minutes Using Massively Parallel Deep
               Reinforcement Learning},
  author    = {Rudin, Nikita and Hoeller, David and Reist, Philipp and
               Hutter, Marco},
  booktitle = {Conference on Robot Learning},
  year      = {2022}
}

@article{li2025generative,
  title={Generative manufacturing systems},
  author={Li, Xingyu and Nassehi, A and Yang, H and Tao, F and Sutherland, J and Wang, L and Gao, R},
  journal={SSRN Electron. J},
  year={2025}
}

@article{li2024diffusion,
  title={Diffusion model for data-driven black-box optimization},
  author={Li, Zihao and Yuan, Hui and Huang, Kaixuan and Ni, Chengzhuo and Ye, Yinyu and Chen, Minshuo and Wang, Mengdi},
  journal={arXiv preprint arXiv:2403.13219},
  year={2024}
}

@article{wu2024diff,
  title={Diff-BBO: Diffusion-Based Inverse Modeling for Black-Box Optimization},
  author={Wu, Dongxia and Kuang, Nikki Lijing and Niu, Ruijia and Ma, Yi-An and Yu, Rose},
  journal={arXiv preprint arXiv:2407.00610},
  year={2024}
}

@article{abgaryan2024llms,
  title={Llms can schedule},
  author={Abgaryan, Henrik and Harutyunyan, Ararat and Cazenave, Tristan},
  journal={arXiv preprint arXiv:2408.06993},
  year={2024}
}

@article{wang2025multi,
  title={Multi-agent large language models as evolutionary optimizers for scheduling optimization},
  author={Wang, Yidan and Wang, Jiayin and Chu, Zhiwei},
  journal={Computers \& Industrial Engineering},
  volume={206},
  pages={111197},
  year={2025},
  publisher={Elsevier}
}

@book{lenstra1977complexity,
  author    = {Lenstra, Jan Karel and Rinnooy Kan, Alexander H. G.},
  title     = {Complexity of Scheduling under Precedence Constraints},
  journal   = {Operations Research},
  volume    = {26},
  number    = {1},
  pages     = {22--35},
  year      = {1977},
  publisher = {INFORMS}
}

@article{brandimarte1993routing,
  author    = {Brandimarte, Paolo},
  title     = {Routing and Scheduling in a Flexible Job Shop by Tabu Search},
  journal   = {Annals of Operations Research},
  volume    = {41},
  number    = {3},
  pages     = {157--183},
  year      = {1993},
  publisher = {Springer}
}

@book{ehrgott2005multicriteria,
  author    = {Ehrgott, Matthias},
  title     = {Multicriteria Optimization},
  year      = {2005},
  edition   = {2nd},
  publisher = {Springer},
  address   = {Berlin, Heidelberg}
}

@article{liaw1998iterative,
  title={An iterative improvement approach for the nonpreemptive open shop scheduling problem},
  author={Liaw, Ching-Fang},
  journal={European Journal of Operational Research},
  volume={111},
  number={3},
  pages={509--517},
  year={1998},
  publisher={Elsevier}
}

@article{zhang2019review,
  title={Review of job shop scheduling research and its new perspectives under Industry 4.0},
  author={Zhang, Jian and Ding, Guofu and Zou, Yisheng and Qin, Shengfeng and Fu, Jianlin},
  journal={Journal of intelligent manufacturing},
  volume={30},
  number={4},
  pages={1809--1830},
  year={2019},
  publisher={Springer}
}

@article{sels2012comparison,
  title={A comparison of priority rules for the job shop scheduling problem under different flow time-and tardiness-related objective functions},
  author={Sels, Veronique and Gheysen, Nele and Vanhoucke, Mario},
  journal={International Journal of Production Research},
  volume={50},
  number={15},
  pages={4255--4270},
  year={2012},
  publisher={Taylor \& Francis}
}

@article{yu2026automated,
  title={Automated Scheduling Heuristic Generation and Evaluation via Large Language Model},
  author={Yu, Fei and Gao, Liang and Li, Xinyu and Lu, Chao and Liu, Qihao},
  journal={IEEE Transactions on Evolutionary Computation},
  year={2026},
  publisher={IEEE}
}

@article{ccetinkaya2025discovering,
  title={Discovering heuristics with Large Language Models (LLMs) for mixed-integer programs: Single-machine scheduling},
  author={{\c{C}}etinkaya, {\.I}brahim O{\u{g}}uz and B{\"u}y{\"u}ktahtak{\i}n, {\.I} Esra and Shojaee, Parshin and Reddy, Chandan K},
  journal={Computers \& Operations Research},
  pages={107325},
  year={2025},
  publisher={Elsevier}
}

@article{liang2024diffsg,
  title={DiffSG: A generative solver for network optimization with diffusion model},
  author={Liang, Ruihuai and Yang, Bo and Yu, Zhiwen and Guo, Bin and Cao, Xuelin and Debbah, M{\'e}rouane and Poor, H Vincent and Yuen, Chau},
  journal={arXiv preprint arXiv:2408.06701},
  year={2024}
}

@article{li2025large,
  title={A large manufacturing decision model for human-centric decision-making},
  author={Li, Xingyu and Nassehi, Aydin and Hu, S Jack and Joung, Byung Gun and Gao, Robert X},
  journal={CIRP Annals},
  year={2025},
  publisher={Elsevier}
}

@inproceedings{krishnamoorthy2023diffusion,
  title={Diffusion models for black-box optimization},
  author={Krishnamoorthy, Siddarth and Mashkaria, Satvik Mehul and Grover, Aditya},
  booktitle={International Conference on Machine Learning},
  pages={17842--17857},
  year={2023},
  organization={PMLR}
}

@inproceedings{li2025expensive,
  title={Expensive multi-objective bayesian optimization based on diffusion models},
  author={Li, Bingdong and Di, Zixiang and Lu, Yongfan and Qian, Hong and Wang, Feng and Yang, Peng and Tang, Ke and Zhou, Aimin},
  booktitle={Proceedings of the AAAI Conference on Artificial Intelligence},
  volume={39},
  number={25},
  pages={27063--27071},
  year={2025}
}

@article{li2024fast,
  title={Fast t2t: Optimization consistency speeds up diffusion-based training-to-testing solving for combinatorial optimization},
  author={Li, Yang and Guo, Jinpei and Wang, Runzhong and Zha, Hongyuan and Yan, Junchi},
  journal={Advances in Neural Information Processing Systems},
  volume={37},
  pages={30179--30206},
  year={2024}
}

@article{ho2022classifier,
  title={Classifier-free diffusion guidance},
  author={Ho, Jonathan and Salimans, Tim},
  journal={arXiv preprint arXiv:2207.12598},
  year={2022}
}

@article{wang2025new,
  title={A new prediction strategy for dynamic multi-objective optimization using diffusion model},
  author={Wang, Feng and Xie, Jinsong and Zhou, Aimin and Tang, Ke},
  journal={IEEE Transactions on Evolutionary Computation},
  year={2025},
  publisher={IEEE}
}

@article{li2026diffusion,
  title={Diffusion Learning-guided Evolution for Large-scale Dynamic Multi-Objective Optimization},
  author={Li, Chenyang and Yen, Gary G and He, Zhenan},
  journal={IEEE Transactions on Evolutionary Computation},
  year={2026},
  publisher={IEEE}
}

@inproceedings{paulus2021comboptnet,
  title={Comboptnet: Fit the right np-hard problem by learning integer programming constraints},
  author={Paulus, Anselm and Rol{\'\i}nek, Michal and Musil, V{\'\i}t and Amos, Brandon and Martius, Georg},
  booktitle={International Conference on Machine Learning},
  pages={8443--8453},
  year={2021},
  organization={PMLR}
}

@inproceedings{ho2020denoising,
  title     = {Denoising Diffusion Probabilistic Models},
  author    = {Ho, Jonathan and Jain, Ajay and Abbeel, Pieter},
  booktitle = {Advances in Neural Information Processing Systems},
  volume    = {33},
  pages     = {6840--6851},
  year      = {2020}
}

@article{hornik1989multilayer,
  title     = {Multilayer feedforward networks are universal approximators},
  author    = {Hornik, Kurt and Stinchcombe, Maxwell and White, Halbert},
  journal   = {Neural Networks},
  volume    = {2},
  number    = {5},
  pages     = {359--366},
  year      = {1989}
}

@inproceedings{schlichtkrull2018modeling,
  title     = {Modeling Relational Data with Graph Convolutional Networks},
  author    = {Schlichtkrull, Michael and Kipf, Thomas N and Bloem, Peter and 
               van den Berg, Rianne and Titov, Ivan and Welling, Max},
  booktitle = {European Semantic Web Conference},
  pages     = {593--607},
  year      = {2018}
}

@inproceedings{dhariwal2021diffusion,
  title     = {Diffusion Models Beat {GAN}s on Image Synthesis},
  author    = {Dhariwal, Prafulla and Nichol, Alexander},
  booktitle = {Advances in Neural Information Processing Systems},
  volume    = {34},
  pages     = {8780--8794},
  year      = {2021}
}

@inproceedings{song2021denoising,
  title     = {Denoising Diffusion Implicit Models},
  author    = {Song, Jiaming and Meng, Chenlin and Ermon, Stefano},
  booktitle = {International Conference on Learning Representations},
  year      = {2021}
}

@article{rahwan2019society,
  title   = {Machine behaviour},
  author  = {Rahwan, Iyad and Cebrian, Manuel and Obradovich, Nick and 
             Bongard, Josh and Dennett, Daniel C and Doyle, Joanna and 
             Driessche, Simon and Dyer, Jeffrey and Epstein, Ziv and 
             Graells-Garrido, Eduardo and others},
  journal = {Nature},
  volume  = {568},
  number  = {7753},
  pages   = {477--486},
  year    = {2019},
  publisher = {Nature Publishing Group}
}

@inproceedings{hu2024multistage,
  title     = {Multi-Stage Predict+Optimize for (Mixed Integer) Linear Programs},
  author    = {Hu, Xinyi and Lee, Jasper C.H. and Lee, Jimmy H.M. and Stuckey, Peter J.},
  booktitle = {Advances in Neural Information Processing Systems},
  volume    = {37},
  year      = {2024},
  url       = {https://proceedings.neurips.cc/paper_files/paper/2024/hash/7718914dfe7d5a657bf6261b5f431021-Abstract-Conference.html}
}

@article{deb2002fast,
  title     = {A Fast and Elitist Multiobjective Genetic Algorithm: {NSGA-II}},
  author    = {Deb, Kalyanmoy and Pratap, Amrit and Agarwal, Sameer and 
               Meyarivan, T},
  journal   = {IEEE Transactions on Evolutionary Computation},
  volume    = {6},
  number    = {2},
  pages     = {182--197},
  year      = {2002}
}

@article{li2008multiobjective,
  title     = {Multiobjective Optimization Problems with Complicated Pareto Sets, 
               {MOEA/D} and {NSGA-II}},
  author    = {Li, Hui and Zhang, Qingfu},
  journal   = {IEEE Transactions on Evolutionary Computation},
  volume    = {13},
  number    = {2},
  pages     = {284--302},
  year      = {2009}
}

@book{pinedo2016scheduling,
  title     = {Scheduling: Theory, Algorithms, and Systems},
  author    = {Pinedo, Michael L},
  edition   = {5th},
  publisher = {Springer},
  year      = {2016}
}

@article{branke2001evolutionary,
  title     = {Evolutionary Optimization in Dynamic Environments},
  author    = {Branke, J{\"u}rgen},
  journal   = {Kluwer Academic Publishers},
  year      = {2001}
}

@article{bengio2021machine,
  title   = {Machine Learning for Combinatorial Optimization: A Methodological 
             Tour d'Horizon},
  author  = {Bengio, Yoshua and Lodi, Andrea and Prouvost, Antoine},
  journal = {European Journal of Operational Research},
  volume  = {290},
  number  = {2},
  pages   = {405--421},
  year    = {2021}
}

@inproceedings{veseli2022learning,
  title     = {Learning to Solve Combinatorial Optimization Problems on 
               Real-World Graphs in Linear Time},
  author    = {Vu\v{c}kovi\'{c}, Dra\v{z}en and others},
  booktitle = {Advances in Neural Information Processing Systems},
  year      = {2022}
}

@inproceedings{kool2018attention,
  title     = {Attention, Learn to Solve Routing Problems!},
  author    = {Kool, Wouter and van Hoof, Herke and Welling, Max},
  booktitle = {International Conference on Learning Representations},
  year      = {2019}
}

@inproceedings{vinyals2015pointer,
  title     = {Pointer Networks},
  author    = {Vinyals, Oriol and Fortunato, Meire and Jaitly, Navdeep},
  booktitle = {Advances in Neural Information Processing Systems},
  volume    = {28},
  year      = {2015}
}

@inproceedings{nazari2018reinforcement,
  title     = {Reinforcement Learning for Solving the Vehicle Routing Problem},
  author    = {Nazari, Mohammadreza and Oroojlooy, Afshin and 
               Snyder, Lawrence and Tak{\'a}c, Martin},
  booktitle = {Advances in Neural Information Processing Systems},
  volume    = {31},
  year      = {2018}
}

@article{zhang2007moea,
  title   = {{MOEA/D}: A Multiobjective Evolutionary Algorithm Based on 
             Decomposition},
  author  = {Zhang, Qingfu and Li, Hui},
  journal = {IEEE Transactions on Evolutionary Computation},
  volume  = {11},
  number  = {6},
  pages   = {712--731},
  year    = {2007}
}

@book{talbi2009metaheuristics,
  title     = {Metaheuristics: From Design to Implementation},
  author    = {Talbi, El-Ghazali},
  publisher = {John Wiley \& Sons},
  year      = {2009}
}

@article{lust2012multiobjective,
  title   = {The Multiobjective Traveling Salesman Problem: A Survey and 
             a New Approach},
  author  = {Lust, Thibaut and Teghem, Jacques},
  journal = {Advances in Multi-Objective Nature Inspired Computing},
  pages   = {119--141},
  year    = {2010}
}

@article{corsini2024self,
  title={Self-labeling the job shop scheduling problem},
  author={Corsini, Andrea and Porrello, Angelo and Calderara, Simone and Dell'Amico, Mauro},
  journal={Advances in Neural Information Processing Systems},
  volume={37},
  pages={105528--105551},
  year={2024}
}

@article{miller1956magical,
  title   = {The Magical Number Seven, Plus or Minus Two: Some Limits on 
             Our Capacity for Processing Information},
  author  = {Miller, George A},
  journal = {Psychological Review},
  volume  = {63},
  number  = {2},
  pages   = {81--97},
  year    = {1956}
}

@inproceedings{lindauer2022smac3,
  title     = {{SMAC3}: A Versatile Bayesian Optimization Package for 
               Hyperparameter Optimization},
  author    = {Lindauer, Marius and Eggensperger, Katharina and Feurer, 
               Matthias and Biedenkapp, Andr{\'e} and Deng, Difan and 
               Sass, Carolin and Bergman, Ed and Hutter, Frank},
  booktitle = {Journal of Machine Learning Research},
  volume    = {23},
  pages     = {1--9},
  year      = {2022}
}

@article{wierzbicki1980use,
  title   = {The Use of Reference Objectives in Multiobjective Optimization},
  author  = {Wierzbicki, Andrzej P},
  journal = {Multiple Criteria Decision Making Theory and Application},
  pages   = {468--486},
  year    = {1980}
}

@inproceedings{sener2018multi,
  title     = {Multi-Task Learning as Multi-Objective Optimization},
  author    = {Sener, Ozan and Koltun, Vladlen},
  booktitle = {Advances in Neural Information Processing Systems},
  volume    = {31},
  year      = {2018}
}

@inproceedings{yang2023foundation,
  title     = {Foundation Models for Decision Making: Problems, Methods, 
               and Opportunities},
  author    = {Yang, Sherry and Nachum, Ofir and Du, Yilun and Wei, Jason 
               and Abbeel, Pieter and Schuurmans, Dale},
  booktitle = {arXiv preprint arXiv:2303.04129},
  year      = {2023}
}

@inproceedings{gilmer2017neural,
  title     = {Neural Message Passing for Quantum Chemistry},
  author    = {Gilmer, Justin and Schütt, Kristof T. and Mayr, Andreas and
               Gastegger, Michael and Marquetand, Philipp},
  booktitle = {International Conference on Machine Learning},
  year      = {2017}
}

@article{battaglia2018relational,
  title   = {Relational Inductive Biases, Deep Learning, and Graph Networks},
  author  = {Battaglia, Peter W. and others},
  journal = {arXiv preprint arXiv:1806.01261},
  year    = {2018}
}

@inproceedings{hamilton2017inductive,
  title     = {Inductive Representation Learning on Large Graphs},
  author    = {Hamilton, William L. and Ying, Rex and Leskovec, Jure},
  booktitle = {Advances in Neural Information Processing Systems},
  year      = {2017}
}

@inproceedings{li2016gated,
  title     = {Gated Graph Sequence Neural Networks},
  author    = {Li, Yujia and Tarlow, Daniel and Brockschmidt, Marc and Zemel, Richard},
  booktitle = {International Conference on Learning Representations},
  year      = {2016}
}

@inproceedings{navon2021learning,
  title     = {Learning the Pareto Front with Hypernetworks},
  author    = {Navon, Aviv and Shamsian, Aviv and Fetaya, Ethan and 
               Chechik, Gal},
  booktitle = {International Conference on Learning Representations},
  year      = {2021}
}

@inproceedings{song2021scorebased,
  title     = {Score-Based Generative Modeling through Stochastic 
               Differential Equations},
  author    = {Song, Yang and Sohl-Dickstein, Jascha and Kingma, 
               Diederik P and Kumar, Abhishek and Ermon, Stefano and 
               Poole, Ben},
  booktitle = {International Conference on Learning Representations},
  year      = {2021}
}

@inproceedings{brown2020language,
  title     = {Language Models are Few-Shot Learners},
  author    = {Brown, Tom B and Mann, Benjamin and Ryder, Nick and 
               Subbiah, Melanie and Kaplan, Jared D and Dhariwal, Prafulla and 
               Neelakantan, Arvind and Shyam, Pranav and Sastry, Girish and 
               Askell, Amanda and Agarwal, Sandhini and Herbert-Voss, Ariel and 
               Krueger, Gretchen and Henighan, Tom and Child, Rewon and 
               Ramesh, Aditya and Ziegler, Daniel M and Wu, Jeffrey and 
               Winter, Clemens and Hesse, Christopher and Chen, Mark and 
               Sigler, Eric and Litwin, Mateusz and Gray, Scott and 
               Chess, Benjamin and Clark, Jack and Berner, Christopher and 
               McCandlish, Sam and Radford, Alec and Sutskever, Ilya and 
               Amodei, Dario},
  booktitle = {Advances in Neural Information Processing Systems},
  volume    = {33},
  pages     = {1877--1901},
  year      = {2020}
}

@inproceedings{touvron2023llama,
  title     = {{LLaMA}: Open and Efficient Foundation Language Models},
  author    = {Touvron, Hugo and Lavril, Thibaut and Izacard, Gautier and 
               Martinet, Xavier and Lachaux, Marie-Anne and Lacroix, Timoth{\'e}e 
               and Rozi{\`e}re, Baptiste and Goyal, Naman and Hambro, Eric and 
               Azhar, Faisal and others},
  booktitle = {arXiv preprint arXiv:2302.13971},
  year      = {2023}
}

@inproceedings{lin2019pareto,
  title     = {Pareto Multi-Task Learning},
  author    = {Lin, Xi and Zhen, Hui-Ling and Li, Zhenhua and 
               Zhang, Qingfu and Kwong, Sam},
  booktitle = {Advances in Neural Information Processing Systems},
  volume    = {32},
  year      = {2019}
}

@inproceedings{fleming2005many,
  title     = {Many-Objective Optimization: An Engineering Design Perspective},
  author    = {Fleming, Peter J and Purshouse, Robin C and Lygoe, Robert J},
  booktitle = {International Conference on Evolutionary Multi-Criterion 
               Optimization},
  pages     = {14--32},
  year      = {2005},
  publisher = {Springer}
}

@article{garey1976complexity,
  title     = {The Complexity of Flowshop and Jobshop Scheduling},
  author    = {Garey, Michael R and Johnson, David S and Sethi, Ravi},
  journal   = {Mathematics of Operations Research},
  volume    = {1},
  number    = {2},
  pages     = {117--129},
  year      = {1976}
}

@misc{amazon2023,
  title        = {Amazon launches a new AI foundation model to power its robotic fleet and deploys its 1 millionth robot},
  author       = {{Amazon}},
  year         = {2024},
  howpublished = {\url{https://www.aboutamazon.com/news/operations/amazon-million-robots-ai-foundation-model}},
  note         = {Accessed: 2025}
}

@inproceedings{hadfield2016cooperative,
  title     = {Cooperative Inverse Reinforcement Learning},
  author    = {Hadfield-Menell, Dylan and Milli, Smitha and Abbeel, Pieter and 
               Russell, Stuart and Dragan, Anca},
  booktitle = {Advances in Neural Information Processing Systems},
  volume    = {29},
  pages     = {3909--3917},
  year      = {2016}
}

@article{li2026generativemanufacturing,
  title   = {Generative Manufacturing},
  author  = {Li, Xingyu and Nassehi, Aydin and Epureanu, Bogdan I. and Cao, Jian and Hu, S. Jack and Wang, Lihui and Gao, Robert X.},
  journal = {Journal of Manufacturing Systems},
  year    = {2026},
  note    = {Accepted for publication, to appear},
  address = {Elsevier}
}

@inproceedings{nichol2021improved,
  title     = {Improved Denoising Diffusion Probabilistic Models},
  author    = {Nichol, Alexander Quinn and Dhariwal, Prafulla},
  booktitle = {Proceedings of the 38th International Conference on Machine Learning},
  pages     = {8162--8171},
  year      = {2021},
  publisher = {PMLR}
}

@inproceedings{watson2022learning,
  title     = {Learning Fast Samplers for Diffusion Models by Differentiating 
               Through Sample Quality},
  author    = {Watson, Daniel and Chan, William and Ho, Jonathan and Norouzi, Mohammad},
  booktitle = {International Conference on Learning Representations},
  year      = {2022}
}

@inproceedings{joshi2019efficient,
  title     = {An Efficient Graph Convolutional Network Technique for 
               the Travelling Salesman Problem},
  author    = {Joshi, Chaitanya K and Laurent, Thomas and Bresson, Xavier},
  booktitle = {arXiv preprint arXiv:1906.01227},
  year      = {2019}
}

@inproceedings{vaswani2017attention,
  title     = {Attention Is All You Need},
  author    = {Vaswani, Ashish and Shazeer, Noam and Parmar, Niki and 
               Uszkoreit, Jakob and Jones, Llion and Gomez, Aidan N and 
               Kaiser, Lukasz and Polosukhin, Illia},
  booktitle = {Advances in Neural Information Processing Systems},
  volume    = {30},
  year      = {2017}
}

@inproceedings{perez2018film,
  title     = {{FiLM}: Visual Reasoning with a General Conditioning Layer},
  author    = {Perez, Ethan and Strub, Florian and de Vries, Harm and 
               Dumoulin, Vincent and Courville, Aaron},
  booktitle = {Proceedings of the AAAI Conference on Artificial Intelligence},
  volume    = {32},
  year      = {2018}
}

@article{hendrycks2016gaussian,
  title     = {Gaussian Error Linear Units ({GELUs})},
  author    = {Hendrycks, Dan and Gimpel, Kevin},
  journal   = {arXiv preprint arXiv:1606.08415},
  year      = {2016}
}

@inproceedings{austin2021structured,
  title     = {Structured Denoising Diffusion Models in Discrete State-Spaces},
  author    = {Austin, Jacob and Johnson, Daniel D and Ho, Jonathan and Tarlow, Daniel and van den Berg, Rianne},
  booktitle = {Advances in Neural Information Processing Systems},
  volume    = {34},
  year      = {2021}
}

@inproceedings{donti2021dc3,
  title     = {{DC3}: A Learning Method for Optimization with Hard Constraints},
  author    = {Donti, Priya L and Rolnick, David and Kolter, J Zico},
  booktitle = {International Conference on Learning Representations},
  year      = {2021}
}

@inproceedings{hottung2022efficient,
  title     = {Efficient Active Search for Combinatorial Optimization Problems},
  author    = {Hottung, Andr{\'e} and Kwon, Yeong-Dae and Murphy, Kevin},
  booktitle = {International Conference on Learning Representations},
  year      = {2022}
}

@inproceedings{li2021learning,
  title     = {Learning to Delegate for Large-Scale Vehicle Routing},
  author    = {Li, Sirui and Yan, Zhongxia and Wu, Cathy},
  booktitle = {Advances in Neural Information Processing Systems},
  volume    = {34},
  year      = {2021}
}

\end{document}